\documentclass{article}
\usepackage{stywhispers,amsmath,epsfig}
\usepackage{algorithmic}
\usepackage{algorithm}
\usepackage{amsfonts}
\usepackage{amssymb}
\usepackage{textcomp}
\usepackage[]{subcaption}

\def\x{{\mathbf x}}

\DeclareMathOperator*{\argmax}{argmax}

\def\x{\mathbf{x}}

\def\e{\mathbf{e}}

\def\s{\mathbf{s}}

\def\E{\mathbf{E}}

\def\bmu{{\boldsymbol\mu}}

\def\bSigma{{\boldsymbol\Sigma}}
\def\hbmu{\hat{\bmu}}
\def\hbSigma{\hat{\bSigma}}

\floatstyle{ruled}
\newfloat{algorithm}{tbp}{loa}
\floatname{algorithm}{Algorithm}

\title{Estimating Target Signatures with Diverse Density}
\twoauthors
 {T. Glenn\sthanks{T. Glenn: tcg@precisionsilver.com}}
	{Precision Silver, LLC\\
	Columbia, MO 65203 USA}
 {A. Zare\sthanks{A. Zare: zarea@missouri.edu, This material is based upon work supported by the National Science Foundation under Grant No. IIS-1350078 - CAREER: Supervised Learning for Incomplete and Uncertain Data.}}
	{University of Missouri\\
	Electrical and Computer Engineering\\
	Columbia, MO 65211 USA}
\begin{document}
%
\maketitle
\begin{abstract}
Hyperspectral target detection algorithms rely on knowing the desired target signature in advance.  However, obtaining an effective target signature can be difficult; signatures obtained from laboratory measurements or hand-spectrometers in the field may not transfer to airborne imagery effectively.  One approach to dealing with this difficulty is to learn an effective target signature from training data.  An approach for learning target signatures from training data is presented.  The proposed approach addresses uncertainty and imprecision in groundtruth in the training data using a multiple instance learning, diverse density (DD) based objective function.  After learning the target signature given data with uncertain and imprecise groundtruth, target detection can be applied on test data.  Results are shown on simulated and real data. 

\end{abstract}
\section{Introduction}
\label{sec:intro}

Many methods for full- and sub-pixel target detection have been developed in the hyperspectral literature \cite{manolakis:2003,Glenn:2013}.  All of these target detection methods rely on having knowledge of the desired target signature in advance.  However obtaining this target signature can be challenging.  For example, laboratory and hand-spectrometer measurements may not be applicable to large hyperspectral scenes due to differences in the measurement characteristics, variations due to environment or atmospheric changes.  Also, in some applications, an analyst may identify targets or regions of interest in a hyperspectral image and wish to identify this target in future data collections or other sets of imagery.  In this later case, the analyst lacks any reference target spectra and, depending on the spatial resolution of the imagery, may lack pure pixels of the target or even precise locations of pixels containing the target. The goal of this paper is to present a hyperspectral target estimation approach aimed at individuals with only approximate knowledge of the locations of sub-pixels targets of interest in an image.  Using a multiple instance learning approach, a discriminative target spectrum is estimated.  

Multiple instance learning is a type of supervised learning in which training data points are not individually labeled \cite{Zare:2014whispers,maron1998framework}.  Instead, sets, or \emph{bags}, containing a variable number of data points are labeled.  A previous method for target spectrum estimation given a multiple instance learning framework was shown in \cite{Zare:2014whispers} where target estimation was conducted used a multiple instance learning-based unmixing method.  The proposed method does not rely on unmixing. Instead a method that optimizes a matched filter output is presented.

In the following, Section \ref{sec:mi1} introduces the DD based objective function that is optimized to learn the needed target signature from training data and describes an evolutionary algorithm to optimize the proposed objective function.  Section \ref{sec:results} shows results on simulated and real hyperspectral data.  Section \ref{sec:summ} provides a summary and description of future work.  

\section{Multiple Instance Hyperspectral Target Estimation}

In the case of hyperspectral target spectral estimation given here, each data point (i.e., pixel in a hyperspectral image) is considered to be a mixture of the pure spectra of the materials found in that pixel's field of view, 
\begin{equation}
\mathbf{x}_i = f(\mathbf{E}_i, \mathbf{p}_i) \text{ where } p_{im} \ge 0 \forall i,m,
\end{equation}
 $\mathbf{x}_i$ is the $i^{th}$ data point,  $\mathbf{E}$ is the set of endmembers (i.e., pure spectral signatures) of the materials found in the full data set, $\mathbf{p}_i$ is the vector of material abundances for pixel $i$ given each endmember in $\mathbf{E}$, $p_{im}$ is the abundance of pixel $i$ for endmember $m$, and $f$ is a function defining the mixture model appropriate for the $i^{th}$ data point.  Consider a training data set partitioned into $K$ bags,  $\mathbf{B} = \left\{ B_1, \ldots, B_K\right\}$with associated bag-level labels, $L = \left\{L_1, \ldots, L_K\right\}$ where $L_j=1$ (labeled positive) if any of the data points in bag $B_j$ have a non-zero abundance associated with the target endmember, $\mathbf{e}_T$, 
\begin{equation}
L_j=1, \text{if }\exists \mathbf{x}_i \in B_j\text{ s.t. } p_{iT} > 0.
\label{eq:l1}
\end{equation}
$L_j=0$ if all data in $B_j$ have zero target proportion, 
\begin{equation}
L_j=0, \text{if } p_{iT} = 0  \forall \mathbf{x}_i \in B_j.
\end{equation}
Since the number of target points (and which points correspond to target points) is unknown, the bag-level labels represent uncertainty in the groundtruth.  In application, this could be related to uncertain groundtruth (e.g., error in GPS coordinates or relying on some general region of interest).  Furthermore, these bag-level labels are imprecise as they only provide a binary indication of whether some subpixel proportion of target can be found in the bag instead of providing indication of the exact abundance amounts.

\label{sec:mi1}
The general definition of DD \cite{maron1998framework} is shown in \eqref{eq:dd},
\begin{equation}
\label{eq:dd}
\argmax_{\mathbf{x}} \prod_{j=1}^{N_p} Pr(\mathbf{x} = \mathbf{e}_T | B_j^+)\prod_{j=1}^{N_n} Pr(\mathbf{x} = \mathbf{e}_T | B_j^-) .
\end{equation}
The terms in \eqref{eq:dd} are often defined using the noisy-or model, 
\begin{eqnarray}
\argmax_{\mathbf{x}} &&\prod_{j=1}^{N_p} 1 - \left( \prod_{i=1}^{N_{pj}} \left( 1 - Pr(\mathbf{x} = \mathbf{e}_T | B_{ji}^+) \right) \right)\nonumber\\ 
&&\prod_{j=1}^{N_n}  \prod_{j=1}^{N_{nj}} \left( 1 - Pr(\mathbf{x} = \mathbf{e}_T | B_{ji}^-) \right).
\label{eq:noisyor}
\end{eqnarray}
The first term in \eqref{eq:noisyor} can be interpreted as enforcing that there is at least one data point in each positive bag containing the target material, $\mathbf{e}_T$.  
Conversely, the second term in \eqref{eq:noisyor}  can be interpreted as saying that there is no target material in any of the points in negative bags.  

The noisy-or model has several limitations that cause difficulties in practice. The first limitation is the product based formulation, which if implemented directly quickly leads to numerical underflow. The common solution here is to use the logarithm of the objective. Secondly, the noisy-or relies upon discrete probabilities and not probability densities, which are often greater than one and incompatible with the formulation. Thirdly, the formulation weights all positive and negative bags equally; weighting parameters are often needed to adjust the relative impact of the terms.

An initial, direct, approach to applying the log of the noisy-or was investigated which used a sigmoid function over a target detector. This approach, however, introduced tuning parameters upon which the model was found to be highly dependent. Also, the inputs to the log terms were often very near one or zero, causing large order of magnitude swings in the range of outputs which are difficult to properly weight among terms. Therefore, a more amenable approach to the DD is proposed that lacks the contortions needed for the direct approach. This general objective \eqref{eq:obj_general} is conceptually based upon the logarithm of an ``or'' model, but uses a different formulation than the noisy-or.
\begin{equation}
  \argmax_{\x} \, \alpha \sum_{j=1}^{N_p} \max_{i} f(\x,B_{ji}^+) + \beta \sum_{j=1}^{N_n} g(\x,B_{j}^-)
  \label{eq:obj_general}
\end{equation}

Our application of \eqref{eq:obj_general} weights the terms such that the mean over all positive bags and negative bags is taken, $\alpha=1/N_{p}$ and $\beta=1/N_{n}$. The $f(\x,B_{ji}^+)$ is taken to be the spectral matched filter detection output of a point in a positive bag given the proposed target signature, 
\begin{equation}
    f(\x,B_{ji}^+) = \frac{(\x - \hbmu)^T \hbSigma^{-1} (B_{ji}^+ - \hbmu)}{\left((\x - \hbmu)^T \hbSigma^{-1} (\x - \hbmu)\right)^{1/2}} 
    \label{eq:obj_pos}  
\end{equation}
where $\hbmu$ and $\hbSigma$ are the mean and covariance estimated from the entire image (for simplicity in this case). The $g(\x,B_{ji}^-)$ is then taken to be the negative of the average detection output with the proposed target signature over all pixels in the negative bags,
\begin{equation}
    g(\x,B_{j}^-) = - \frac{1}{N_{nj}} \sum_{i=1}^{N_{nj}} \frac{(\x - \hbmu)^T \hbSigma^{-1} (B_{ji}^- - \hbmu)}{\left((\x - \hbmu)^T \hbSigma^{-1} (\x - \hbmu)\right)^{1/2}} .  
    \label{eq:obj_neg}
\end{equation}
We note that this objective formulation is applicable to any choice of detection statistic with a bigger-is-better output. We also acknowledge that this does not have a direct probabilistic interpretation, but we opted for simplicity in this case.

To optimize the DD function with respect to the target signature, an evolutionary algorithm was used. The evolutionary algorithm optimizes by iteratively mutating and selecting from a population of potential solutions.  The algorithm begins by initializing a $N_{pop}$-sized population of potential solutions, $\E_{pop} =\{ \e_{1},\ldots, \e_{N_{pop}}\}$. In our implementation, the population is initialized by setting one element of the population to the one pixel from all of the positive bags with the largest objective function value (as shown in \ref{eq:obj_general}). The remaining elements are initialized by randomly selecting pixels from the positive bags.  After initialization, the population is mutated to generate a child population, $\E_{pop}^{\prime} =\{ \e_{1}^{\prime},\ldots, \e_{N_{pop}}^{\prime}\}$. The mutation is conducted by randomly selecting a wavelength to mutate and, then, adding random noise to that wavelength of the parent solution. The added noise is generated according to a two-component zero-mean Gaussian mixture, $r \sim w_nN(\cdot|0,\sigma_n) + (1-w_n)N(\cdot|0,\sigma_w)$, where $w_n \in [0, 1]$ and $\sigma_n << \sigma_w$ to allow for both ``small'' and ``large'' mutations. After mutation, the parent and children solutions are pooled into a single set and the $N_{pop}$ solutions with largest objective function values are maintained for the next iteration. The mutation-selection process is repeated for $N_{iter}$ iterations. The solution with the largest objective function value in the final iteration is returned as the estimated target. 


\begin{figure*}[htb]
\begin{subfigure}[]{.5\textwidth}
\centering
\centerline{\epsfig{figure=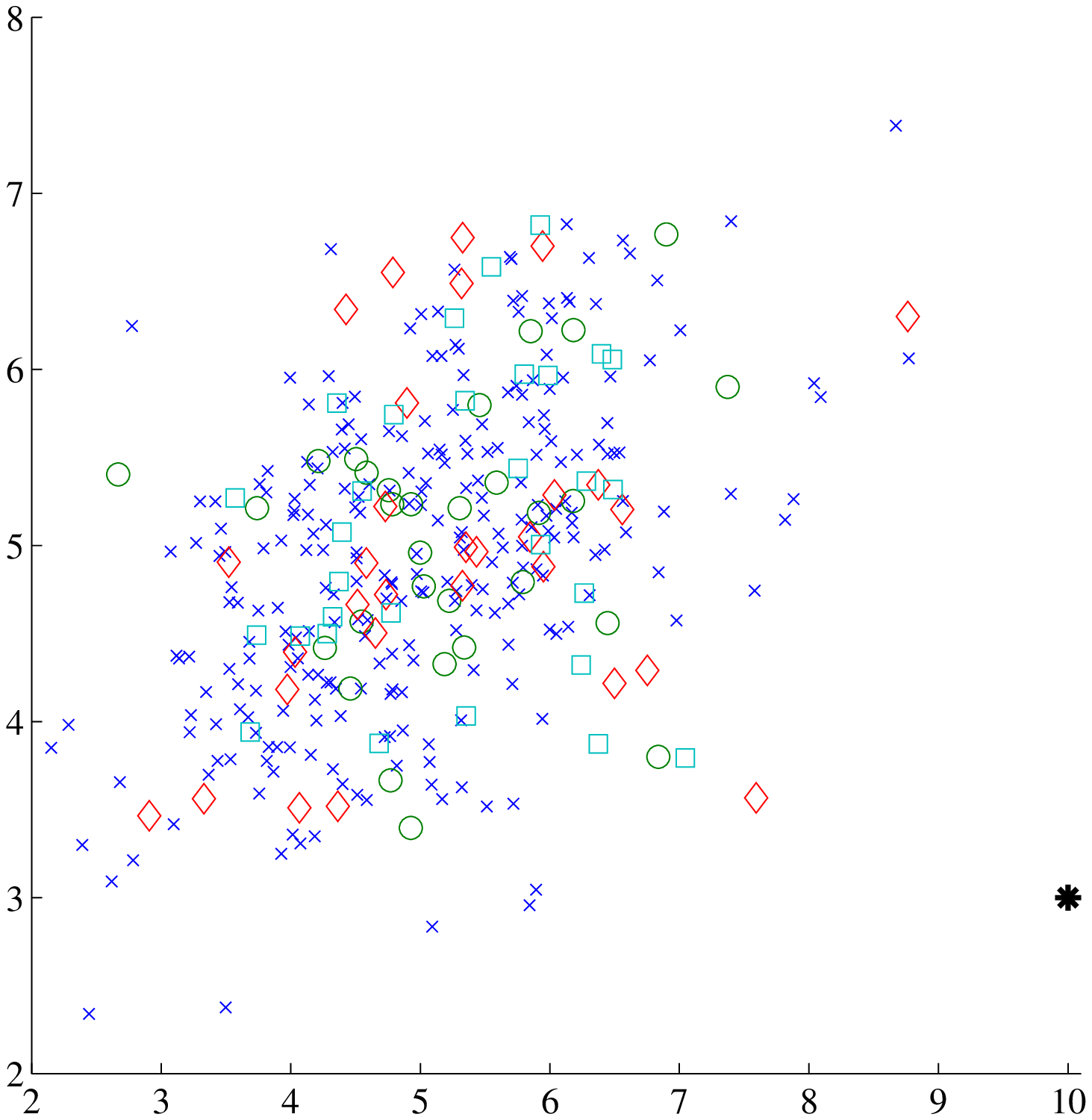,width=5.0cm}}
\caption{}
\label{fig:synth_train_scatter}
\end{subfigure}
\begin{subfigure}[]{.5\textwidth}
\centering
\centerline{\epsfig{figure=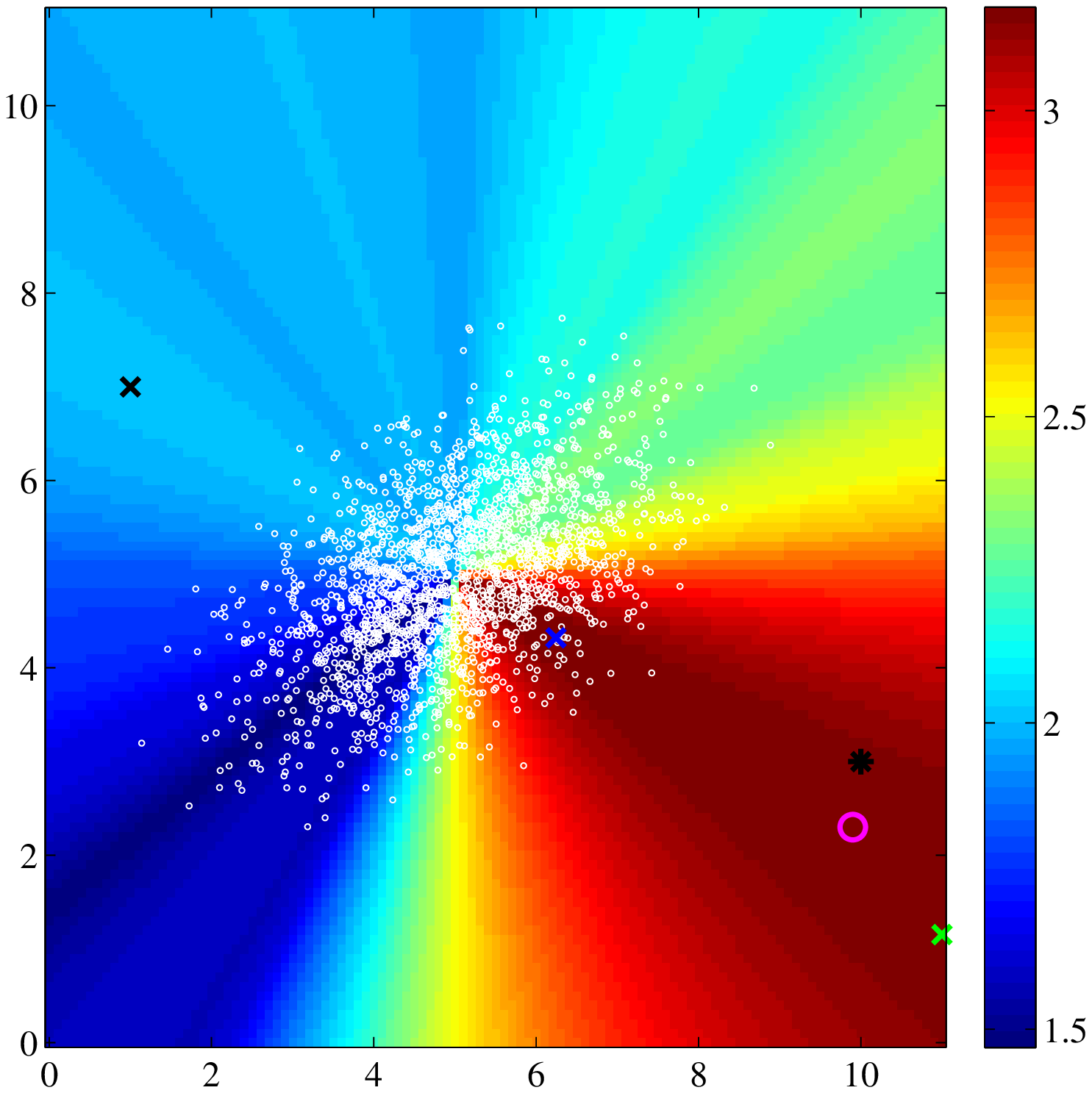,width=5.0cm}}
\caption{ }
\label{fig:synth_exp}
\end{subfigure}
\caption{(a) Synthetic Data Training Set. Blue X's are negative bag pixels. Green circles, red diamonds, and cyan squares are pixels from each of 3 positive bags. Black star is true target signature for reference. (b) Synthetic Data Results. Background color shows the DD objective function value for that point in feature space. White points are a subset of pixel values. Black X at $(1,7)$ is purposefully poor initialization. Blue X at approx $(6,4)$ is positive bag pixel with best objective value. Green X at approx $(11,1)$ is learned signature. Magenta circle at approx $(10,2.5)$ is the optimum objective value on $0.01$ resolution grid. Black star at $(10,3)$ is the true target value.}
\end{figure*}

\section{Results}
\label{sec:results}

A simulated image dataset was constructed in order to demonstrate the conceptual effectiveness of the proposed method. An image with 100 rows and 100 columns was created with pixel values drawn from a Gaussian distribution with $\bmu = (5, 5)^T$ and a covariance with $\sigma_{11}=\sigma_{22}=1$ and $\sigma_{12}=\sigma_{21}=0.5$. The target signature used was $\s=(10, 3)^T$, and 100 pixels were chosen to include the target on a regular grid of $(\mathrm{row},\mathrm{column})$ indexes (i.e. $(5,5)$, $(5, 15)$, $\ldots$, $(5, 95)$, $(15, 5)$, and so on). Target pixels were mixed with a random proportion of $25\%$ to $50\%$ target. Three target bags of 30 pixels were selected from the image with each containing a single target-mixed pixel. Three negative bags of 80 pixels were also selected, each correspondingly having no target-mixed pixels. Figure~\ref{fig:synth_train_scatter} shows a scatter plot of the training samples used as well as the true target signature.


The proposed optimization algorithm was run with the positive and negative bags selected in order to learn an appropriate target signature. The mean and inverse covariance for the matched filter were estimated from all pixels in the image (including the target-mixed pixels). Additionally, the proposed objective was evaluated using the training positive and negative bags at every multiple of $0.01$ on the 2D grid from $[0,11]$. These results are shown in Figure~\ref{fig:synth_exp}. 

Several important items are shown by this experiment.  First, the objective function given the training data behaves well, having maximal output in approximately the direction of the true target signature. This is done without explicitly specifying which pixels contain the target, and indeed in examining Figure~\ref{fig:synth_train_scatter}, it would be difficult to determine which pixels contain the target or the target signature value were it not shown. Secondly, this shows that the proposed optimization method is able to obtain a good answer, even when deliberately started from a poor location in feature-space which required traversing lower-valued regions to reach the high-value result.

\begin{figure*}[tb!]
\begin{subfigure}[b]{.3\textwidth}
\centering
\centerline{\epsfig{figure=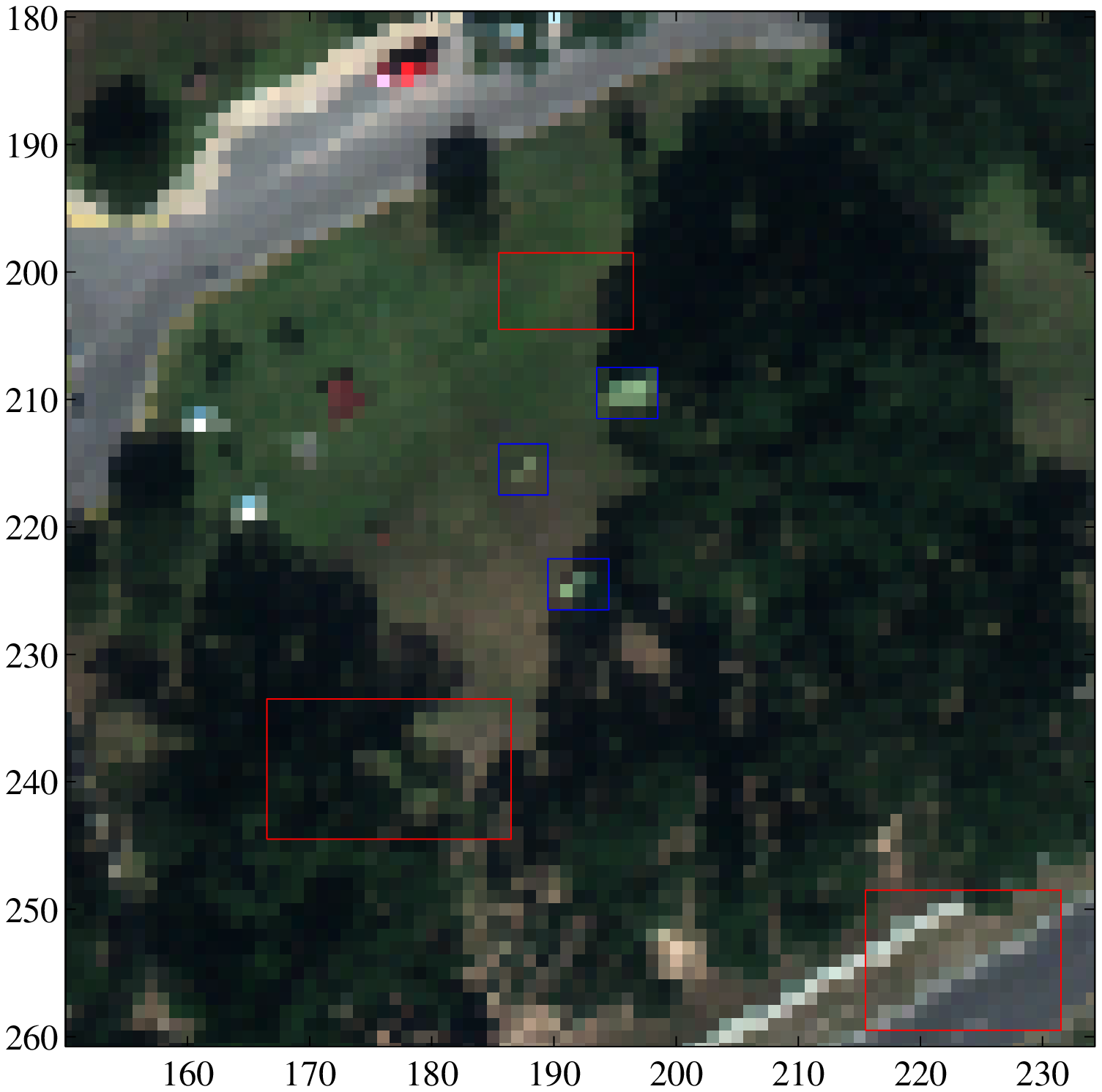,width=4.5cm}}
\caption{}
\label{fig:real_bb_zoom}
\end{subfigure}
\centering
\begin{subfigure}[b]{.3\textwidth}
\epsfig{figure=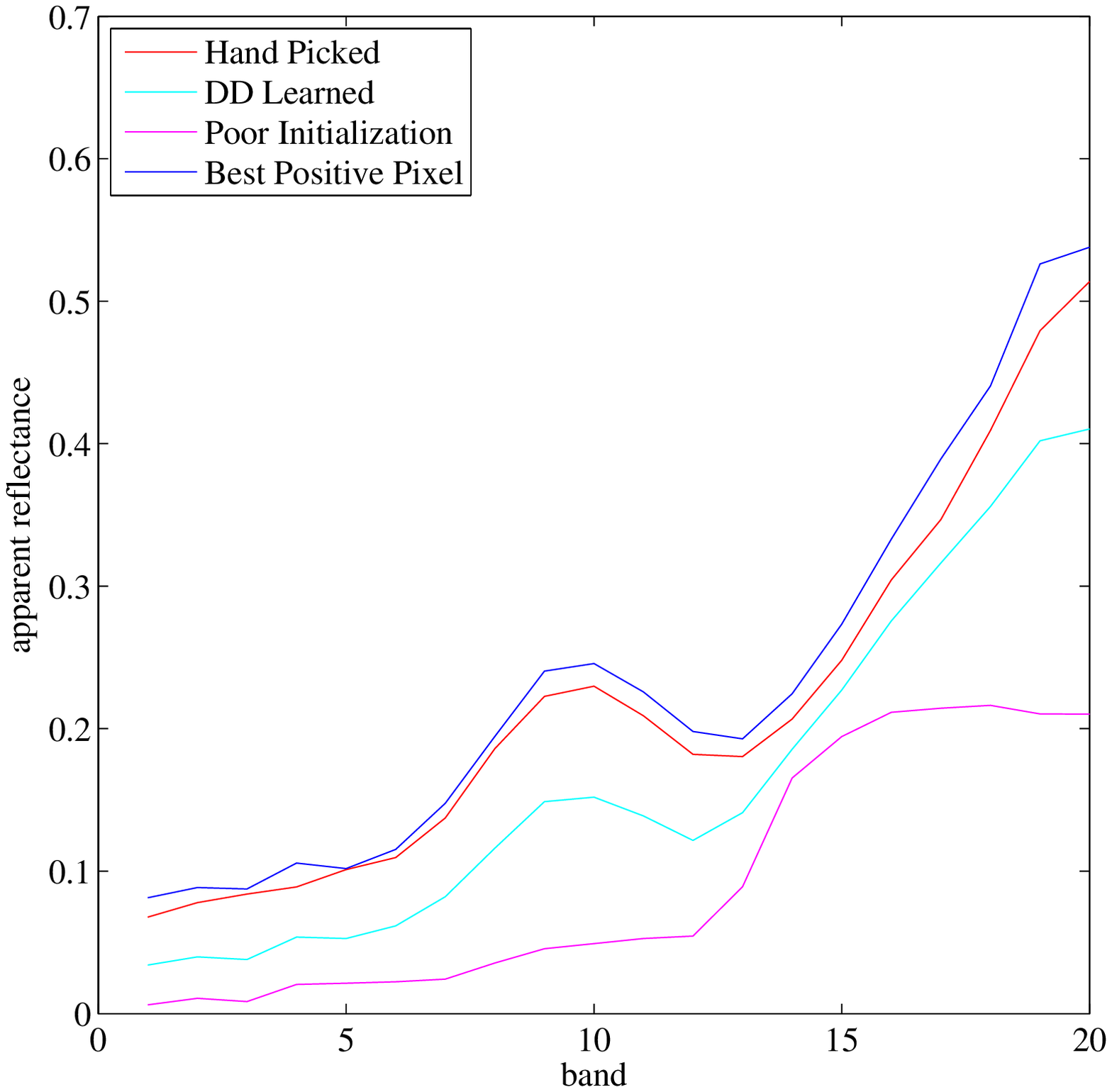,width=4.5cm}
\caption{}
\label{fig:real_spectra}
\end{subfigure}
\begin{subfigure}[b]{.3\textwidth}
\centering
\epsfig{figure=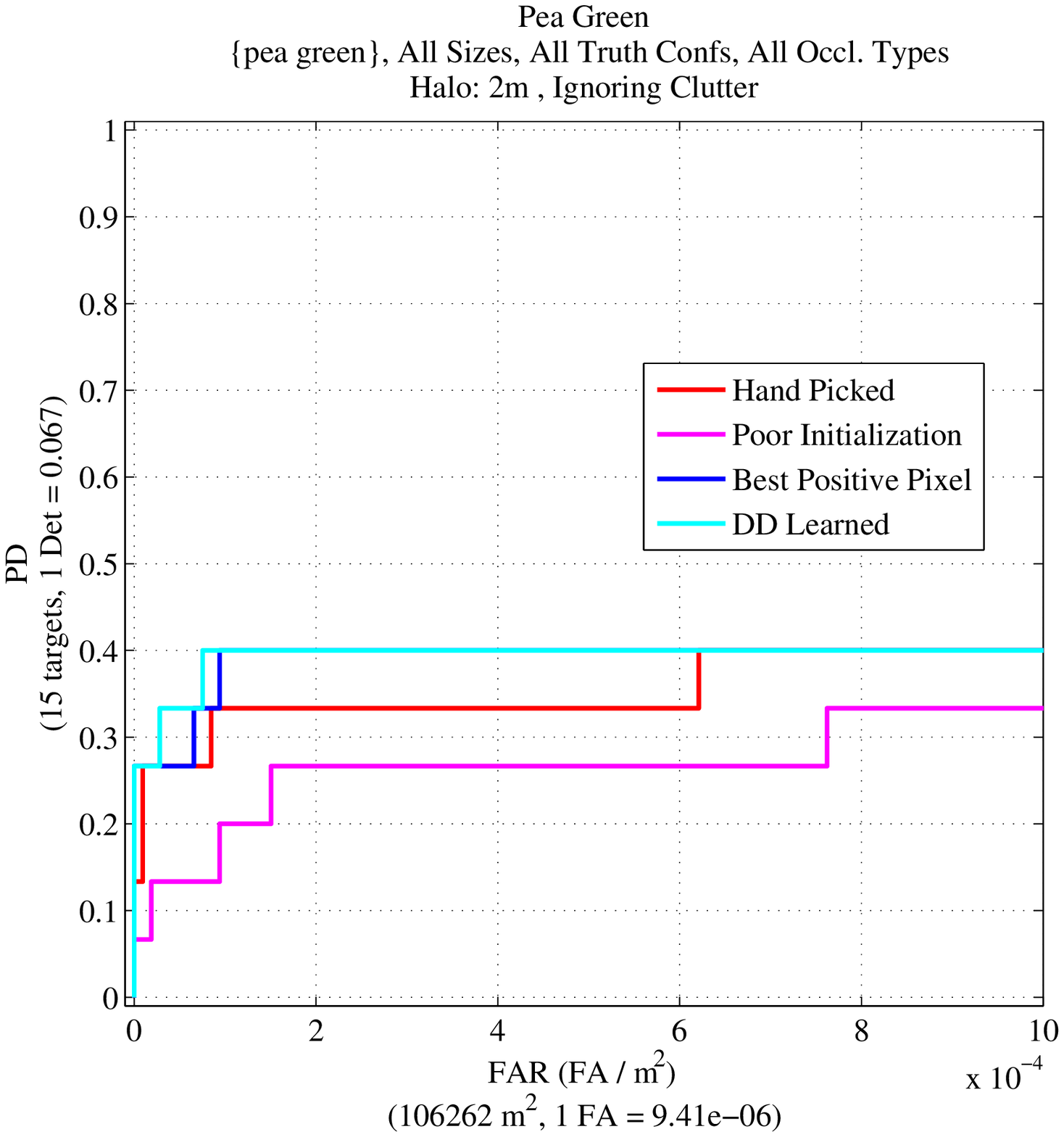,width=4.5cm}
\caption{}
\label{fig:real_rocs}
\end{subfigure}
\caption{(a) Zoomed subimage of MUUFL Gulfport RGB image with positive and negative bag selections. (b) Spectra of DD learned signature, best positive-bag pixel by DD, hand-selected signature, and purposefully poor initialization selection. (c) ROC curves for learned, hand-selected, best positive-bag pixel, and initialization target signatures.}
\end{figure*}

To show the applicability of the method, we tested it with a airborne hyperspectral dataset designed for target detection. We use the MUUFL Gulfport hyperspectral data, collected over the University of Southern Mississippi-Gulf Park campus~\cite{Gader:2013}. Data were collected with an ITRES Inc. hyperspectral Compact Airborne Spectrographic Imager (CASI-1500) over the 375-1050\,nm range with seventy-two 10\,nm spectral bands. The image is $325 \times 337$ pixels in size, and each pixel corresponds to $1\,m^2$. Fifteen examples of four different color cloth targets are emplaced in the scene, with sizes of 0.5m x 0.5m (always sub-pixel), 1m x 1m (probably sub-pixel), and 3m x 3m (at least one pure pixel). Targets were placed in locations of varying shading and occlusion, and many are probably not detectable. As pre-processing, the image was dimensionality reduced to 20 bands using a hierarchical band merging based upon mutual information \cite{Martinez-Uso:2007}.

Figure~\ref{fig:real_bb_zoom} shows a subset of the Gulfport campus image with bounding boxes drawn for the positive and negative bag selections. The positive bag selections, shown in blue, enclose three of the easily visible pea green targets. In total 15 Pea Green targets are present in the full image, many at fully subpixel resolutions. Approximately 6-7 of of the 15 targets are detectable at a reasonable false alarm rate depending upon the detector used. The red boxes are the negative bag selections, which enclose a variety of ground-cover classes like grass, trees, shadow, bare soil, and asphalt, but no targets.

For comparison, a single pixel was hand selected from the pea green targets to use as a comparison target signature. This pixel was selected using the ground truth information to find an unoccluded, fully illuminated, fully resolved pixel from one of the 3m x 3m targets. A few candidate pixels were evaluated based on their detection performance (and known ground-truth), and the best one was selected. 
To test the performance of the DD target estimation, we used the positive and negative bags for the pea green targets to learn a new target signature. In order to show that the proposed method is capable of effectively searching over the high-dimensional space of possible target signatures, we started the search from a purposefully poor initial spectrum computed from the mean of the first negative bag. In real application we recommend initializing to the positive bag pixels that have the maximum DD. Figure~\ref{fig:real_spectra} shows these spectra. 

From Figure~\ref{fig:real_spectra} it can be seen that the best positive bag pixel and the previously hand selected pixel are quite similar. These actually are neighboring pixels from the same target. The DD optimization learns a spectrum with a similar shape to the hand selected spectra as well, though with a lower magnitude. We note that, after mean subtraction, the spectral matched filter effectively normalizes the target signature, and so only depends upon the spectral shape of the target signature and not its magnitude. 

Figure~\ref{fig:real_rocs} shows the resulting target detection performance of each of the spectra using a Receiver Operating Characteristics (ROC) curve up to a False Alarm Rate (FAR) of $10^{-3}$ false alarms per $m^2$. This FAR maximum is approximately 1 false alarm per 30 pixel square region in the image, and thus detections at higher rates are not worth considering and are likely to simply be ``lucky'' detections. These ROC curves show that both the DD learned spectrum, and the best single positive bag pixel by DD are good target signatures. These both provide detection performance that is better than the previous hand-selected spectrum, with the DD learned spectrum being the best by a slim margin. In reality any of these spectra are probably a good prototype choice, and given their similar shapes, the differences in performance are probably attributable to slightly better adaptation of each for discriminability against the ``noise'' characteristics of the background. Importantly the DD learned signature should be better than the best single pixel if the targets were all of subpixel proportion. Most importantly however, this experiment shows that the target spectra can be learned when having only a rough idea of target locations, which is often the case in application.

\section{Summary and Future Work}
\label{sec:summ}
In summary, a multiple instance learning approach for target estimation is presented.  The approach relies on a matched filter target detector with a given background mean and covariance and estimates the most discriminative target signature by maximizing a DD objective function.  

Many extensions to this method are possible, including application with different target detectors and different background estimation methods. More thorough investigation should also be performed to evaluate the effects of the exact form of the objective, the numbers and sizes of the training bags, and to compare with the FUMI methods of signature estimation. 


{\small \bibliographystyle{IEEEbib}
\bibliography{strings,refs}}

\end{document}